\documentclass[]{article}

\usepackage{graphicx}
\usepackage{threeparttable}
\usepackage{mathrsfs,amsmath,amssymb, upgreek}
\usepackage{pbox}
\usepackage{algorithm}
\usepackage{algpseudocode}
\usepackage{multirow}
\usepackage{booktabs}
\usepackage{layouts}
\usepackage[colorlinks=true,linkcolor=blue, citecolor=blue, urlcolor=blue]{hyperref}
\usepackage{etoolbox}
\usepackage{float}
\usepackage[caption = false]{subfig}
\usepackage{bm}
\usepackage{setspace}

\usepackage[noabbrev]{cleveref}
\usepackage{natbib}
\setcitestyle{authoryear}

\newtheorem{definition}{Definition}

\title{How to Estimate the Ability of a Metaheuristic Algorithm to Guide Heuristics During Optimization}
\author{Milo{\v s} Simi{\' c}\footnote{ORCID: orcid.org/0000-0003-1506-3728}\\ University of Belgrade\\  Studentski trg 1, 11000 Belgrade\\ \texttt{milos.simic.csci@gmail.com} }
\date{ }

\usepackage{color}

\begin{document}
	\maketitle
	\begin{abstract}
		Metaheuristics are general methods that guide application of concrete heuristic(s) to problems that are too hard to solve using exact algorithms. However, even though a growing body of literature has been devoted to their statistical evaluation, the approaches proposed so far are able to assess only coupled effects of metaheuristics and heuristics. They do not reveal us anything about how efficient the examined metaheuristic is at guiding its subordinate heuristic(s), nor do they provide us information about how much the heuristic component of the combined algorithm contributes to the overall performance. In this paper, we propose a simple yet effective methodology of doing so by deriving a naive, placebo metaheuristic from the one being studied and comparing the distributions of chosen performance metrics for the two methods. We propose three measures of difference between the two distributions. Those measures, which we call BER values (benefit, equivalence, risk) are based on a preselected threshold of practical significance which represents the minimal difference between two performance scores required for them to be considered practically different. We illustrate usefulness of our methodology on the example of Simulated Annealing, Boolean Satisfiability Problem, and the Flip heuristic.
	\end{abstract}
\section{Introduction}\label{sec:introduction}
Metaheuristics and heuristics are widely accepted optimization tools within operations research community \citep{Caserta2010}. They are used to approximately, but efficiently, solve problems that are too hard to be solved using exact algorithms \citep{Nesmachnow2014}.

Heuristics are problem-specific techniques that, in general, quickly find good solutions to given problems, although there are no guarantees that those solutions will always be optimal. Heuristics can be used only to solve problems for which they have been specifically designed. Metaheuristics, on the other hand, have so far been utilized in two ways \citep{Caserta2010}:
\begin{itemize}
	\item as general-purpose optimization methods ready to apply to any problem without any modification, and
	\item as higher-order methods which guide how problem-specific heuristics are applied to instances belonging to a particular problem class.
\end{itemize}
Over time, it has been noticed that the latter approach yields better results \citep{Caserta2010}. However, once a researcher evaluates such a method, they assess the combined performance of the metaheuristic and its subordinate heuristic(s). Although that gives insight into performance of the method as a whole, it does not provide answers to the following questions:
\begin{itemize}
	\item Is it possible that the performance score has been achieved mostly or solely by the heuristic(s)?
	\item How much does the guiding logic of the metaheuristic contribute to total performance?
\end{itemize} 
The answers to these questions are important because if it is the case that performance comes mostly or solely from heuristics, then it would be wrong to attribute the score to the metaheuristic and claim that a new solver for the specific class of problems has been found. The goal of this paper is to present a sound  methodological framework to answer said questions. To our best knowledge, this is the first attempt to formulate such a technique. 

The rest of the paper is organized as follows. The proposed methodology is described in Section \ref{sec:proposed_methodology}. In Section \ref{sec:example}, we present an example of its application to Simulated Annealing, Boolean Satisfiability Problem and the Flip heuristic. Finally, we discuss it and draw our conclusions in Section \ref{sec:conclusion}.

\section{Proposed Methodology}\label{sec:proposed_methodology}
Let $\mathcal{M}$ be the metaheuristic being examined, and let $\mathcal{H}$ denote a single heuristic or a group of heuristics $\mathcal{H}=\{\mathcal{H}_1, \mathcal{H}_2, \ldots, \mathcal{H}_m\}$ intended to be executed one after another, known to work well on the problem class of interest. The performance metric of the combined method $\mathcal{M}[\mathcal{H}]$ in which $\mathcal{M}$ guides the application of $\mathcal{H}$ can be modeled as a random variable $Y$ whose distribution is given by:
\begin{equation}\label{eq:performance_distribution}
	P(Y | \Pi, S, \theta_{\mathcal{M}}, \theta_{\mathcal{H}})
\end{equation}
where  $\Pi$ and $S$  are random variables representing the instance of the problem class to which $\mathcal{M}[\mathcal{H}]$ is applied and  the seed for the random number generator, whereas $\theta_{\mathcal{M}}$ and $\theta_{\mathcal{H}}$ denote the parameters of $\mathcal{M}$ and $\mathcal{H}$, respectively. For now, we assume that $Y$ is a univariate variable, i.e. that the metric is a single value (the objective function to optimize, execution time, etc.). Its distribution is not known in advance and researchers estimate it by first tuning $\theta_{\mathcal{M}}$ and $\theta_{\mathcal{H}}$ and then evaluating the method on a number of problem instances $\pi_1,\pi_2,\ldots,\pi_l$, repeating evaluation several times for different choices of the seeds for the random number generator.

As said in introduction, the metric $Y$ measured in this manner represents an estimate of the performance of $\mathcal{M}[\mathcal{H}]$. In order to assess how good $\mathcal{M}$ is at guiding $\mathcal{H}$, we can introduce an additional variable $M$ to Equation \ref{eq:performance_distribution} which now becomes:
\begin{equation}\label{eq:performance_distribution_extended}
P(Y | M, \Pi, S, \theta_M, \theta_{\mathcal{H}})
\end{equation}
assuming a more general form for the performance of a metaheuristic ($M$) guiding $\mathcal{H}$ for the problem of a given class. The variable $M$ will denote the metaheuristic component and will be understood to have two levels: $\mathcal{M}$ and $\emptyset$, where the latter denotes what we will call a naive or placebo metaheuristic henceforth. It is a metaheuristic which is based on no purposeful logic and has no components other than random decisions. It is such that $\emptyset[\mathcal{H}]$ acts as an algorithm where $\mathcal{H}$ is guided randomly, as if no metaheuristic has been used to guide it. Then, to answer the question:
\begin{itemize}
	\item How good is $\mathcal{M}$ at guiding $\mathcal{H}$?
\end{itemize}
we should estimate
\begin{equation}
	P(Y | M=\mathcal{M}, \Pi, S, \theta_{\mathcal{M}}, \theta_{\mathcal{H}})
\end{equation}
and compare it to:
\begin{equation}
	P(Y | M=\mathcal{\emptyset}, \Pi, S, \theta_{\emptyset}, \theta_{\mathcal{H}})
\end{equation}
The difference reveals the effect of changing the metaheuristic from naive random search, which has no guiding logic, to $\mathcal{M}$. If the effect is negligible, then it indicates that using $\mathcal{M}$, which may have sophisticated and complicated logic, to guide how $\mathcal{H}$ is applied, is the same as using a naive metaheuristic with no logic to guide $\mathcal{H}$. In fact, that would mean that any score $\mathcal{M}[\mathcal{H}]$ has achieved  comes from using $\mathcal{H}$ and has nothing or little to do with $\mathcal{M}$. After all, if the logic of $\mathcal{M}$ guides $\mathcal{H}$ similarly or identically to random search, then we cannot justify use of $\mathcal{M}$ in that particular setting.

This method is similar to the one used in a typical scenario where there are two factors, $A_1$ and $A_2$, and a researcher wants to estimate the linear effect of $A_1$ on a yield variable $Y$ when $A_2$ is fixed to a certain value. The way to do so is to define the low and high levels of $A_1$ and then estimate how $Y$ changes when $A_1$ is increased from its low to high level. The effect that we are estimating is called the simple effect of $A_1$ at the chosen level of $A_2$. This is precisely what we are trying to do in our case. We want to estimate how the performance metric changes when $M$, the metaheuristic component, is changed from its low level with no logic ($\emptyset$), to its high level, the metaheuristic $\mathcal{M}$ being examined, with the heuristic component fixed to $\mathcal{H}$. The method that we propose in this Section achieves just that. Another example analogous to our case is from pharmacological studies. When a new medicine is tested, one group of patients, called the control group, is given placebo, while the other is given the drug. If the effect of the medicine is significantly better then that of placebo, the drug is deemed effective. Otherwise, there is no justification to produce and use the medicine as it is less effective than a simple placebo. If we use terminology from that example in our study, we will say that the $\emptyset$ acts as placebo and $\mathcal{M}$ takes the role of medicine.

We have exposed the core of our methodology and the rationale behind it, but there are still several issues that we must address:
\begin{enumerate}
	\item Is $\emptyset$ the same for every $\mathcal{M}$ and why cannot we simply apply $\mathcal{H}$ without plugging it in a metaheuristic? 
	\item  Should the parameters be tuned to yield the maximal performance prior to evaluation of $\mathcal{M}[\mathcal{H}]$ and $\emptyset[\mathcal{H}]$ or drawn randomly from a predefined space of allowed values?
	\item How to compare $P(Y | M=\mathcal{M}, \Pi, S, \theta_{\mathcal{M}}, \theta_{\mathcal{H}})$ to $P(Y | M=\mathcal{\emptyset},  \Pi, S, \theta_{\emptyset}, \theta_{\mathcal{H}})$?
\end{enumerate}
We answer all those questions in continuation of this section.

\subsection{The Naive Metaheuristic}
The main idea of our method is to see if guiding $\mathcal{H}$ with no logic is the same as guiding it with the logic of $\mathcal{M}$, the metaheuristic being examined. The rationale behind this is that each metaheuristic is a specific set of rules, and that if using those rules gives the same results as not using any rules at all, then the observed performance is achieved by heuristics alone and the logic of $\mathcal{M}$ is not effective nor efficient. We have referred to  guiding heuristics with no logic as the naive or placebo metaheuristic, $\emptyset$. The reason why, in general, $\emptyset$ has to be a naive metaheuristic, and not just a mere application of $\mathcal{H}$ is that $\emptyset[\mathcal{H}]$ has to invest the same computational effort as $\mathcal{M}[\mathcal{H}]$ in order for comparison of the corresponding distributions to be fair. This means that if $\mathcal{M}$ is a population metaheuristic (such as, e.g., Genetic Algorithm), $\emptyset$ must be too. Similarly, if $M$ is a single solution metaheuristic (such as, e.g., Simulated Annealing), so must be $\emptyset$. Moreover, in the former case, if the population in $\mathcal{M}$ consists of $N$ individuals, the same must hold for $\emptyset$. 

In general, we do not need to state $\emptyset$ explicitly. We can derive $\emptyset[\mathcal{H}]$ from $\mathcal{M}[\mathcal{H}]$ by removing all $\mathcal{M}$'s unique algorithmic components and leaving only naive, random operations. An example in Section \ref{sec:example} will clarify this step.

\subsection{Choice of Parameters}
The choice of parameters is crucial to performance of a (meta)heuristic. If tuned appropriately, they can greatly improve performance. If not set to appropriate values, they can deteriorate the algorithm. The question that naturally arises in our case is whether the parameters should be tuned prior to evaluation or treated as random variables and randomly set before each run of $\mathcal{M}[\mathcal{H}]$ and $\emptyset[\mathcal{H}]$ during their evaluation. Both alternatives are viable, but are related to essentially different goals. If we opt for randomly setting the parameters, we would be aiming to assess the intrinsic guiding capability of $\mathcal{M}$ that does not depend on the choice of parameters and is present in all its applications. However, does such capability exist? Different parameter settings can lead to diametrically opposite results. Besides, before $\mathcal{M}[\mathcal{H}]$ is applied to real problems in practice, it is always tuned. Practitioners and researchers are interested in the best performance $\mathcal{M}[\mathcal{H}]$ can give for a class of problems, not any performance for random parameter settings. Hence, we argue for tuning the parameters of $\mathcal{M}[\mathcal{H}]$ prior to its evaluation. We can use some of them as the parameters of $\emptyset[\mathcal{H}]$ (for example, the number of individuals to ensure the populations are of the same sizes in $\mathcal{M}$ and $\emptyset$), and then tune the parameters of $\mathcal{H}$, if any. 

\subsection{Comparison of Distributions}
In literature, the most common way to compare two metaheuristics is to compare their expected values of $Y$, the chosen performance metric, approximated by means of measurements of $Y$ on the selected problems instances for different, but random choices of the seed for the random number generator. However, we argue against using means to compare distributions of $Y$. What must be understood is that mean, even when accompanied by standard deviation, may not be representative of the distribution \citep{Gunawardena2014}. Therefore, difference in means may not be informative and inference based on it may be invalid. Another, unfortunately common practice that we argue against is using $p$ values as definite proofs to accept or reject tested hypotheses. One reason is that significance at the desired level can always be achieved by using sufficiently large samples \citep{Demidenko2016}. In our case, by evaluating algorithms on a large number of problem instances and repeating the process for a lot of times, we can make $p$ values as small as desired. The other reason is, as \cite{Fraser2016} explained, that "[$p$ value]  can  guide  the  judgments  about  scientific  conclusions, but  cannot  replace  them.".

Knowing this, we ask what is the appropriate way to compare the distribution of the performance metric for $\mathcal{M}[\mathcal{H}]$, $Y_{\mathcal{M}[\mathcal{H}]}$, with that for $\emptyset[\mathcal{H}]$, $Y_{\emptyset[\mathcal{H}]}$? Let us assume, without loss of generality, that lower scores of the metric signify superior performance. If $\mathcal{M}[\mathcal{H}]$ works better than $\emptyset[\mathcal{H}]$, we should expect the distribution of $Y_{\mathcal{M}[\mathcal{H}]}$ to be located to the left of $Y_{\emptyset[\mathcal{H}]}$. A measure of how far the former is to the left of the latter is 
\begin{equation}
P(Y_{\mathcal{M}[\mathcal{H}]} < Y_{\emptyset[\mathcal{H}]})
\end{equation}
the probability that a score of a run of $\mathcal{M}[\mathcal{H}]$ is lower, i.e. better than that of a score of a run of  $\emptyset[\mathcal{H}]$ for some randomly selected problem instance. However, we should not limit ourselves to testing only if $Y_{\mathcal{M}[\mathcal{H}]}$ is located to the left of $Y_{\emptyset[\mathcal{H}]}$. For example, if the scores of $\mathcal{M}[\mathcal{H}]$ lied in the range $(1.1\times10^{-4}, 1.2\times10^{-4})$ and those of $\emptyset[\mathcal{H}]$ in $(1.3\times10^{-4},1.4\times^{-4})$, but score differences lower than $10^{-3}$ are practically negligible, even though $P(Y_{\mathcal{M}[\mathcal{H}]} < Y_{\emptyset[\mathcal{H}]})$ would be equal to $1$ and indicate complete superiority of $Y_{\mathcal{M}[\mathcal{H}}$ over $Y_{\emptyset[\mathcal{H}]}$, which would be true from a purely statistical point of view, but false from the standpoint of practical importance. Therefore, we first need to set some threshold $\delta \geq 0$ to define a minimal difference between two performance scores required for them to be considered practically different. Therefore, instead of estimating $P(Y_{\mathcal{M}[\mathcal{H}]} < Y_{\emptyset[\mathcal{H}]})$, we should focus on:
\begin{equation}
b = P(Y_{\mathcal{M}[\mathcal{H}]} < Y_{\emptyset[\mathcal{H}]} - \delta)
\end{equation}
The probability quantifies practical benefit, with respect to $\delta$, of guiding $\mathcal{H}$ with $\mathcal{M}$, hence the name $b$. The converse probability
\begin{equation}
r=P(Y_{\mathcal{M}[\mathcal{H}]} > Y_{\emptyset[\mathcal{H}]} + \delta)
\end{equation}
represents the risk of using $\mathcal{M}[\mathcal{H}]$ instead of $\emptyset[\mathcal{H}]$, that is the probability that $\mathcal{M}$ guides $\mathcal{H}$ practically worse than $\emptyset$. What remains is the probability that $\mathcal{M}[\mathcal{H}]$ and $\emptyset[\mathcal{H}]$ are practically equivalent:
\begin{equation}
e = P(Y_{\emptyset[\mathcal{H}]} - \delta \leq Y_{\mathcal{M}[\mathcal{H}]} \leq Y_{\emptyset[\mathcal{H}]} + \delta) = 1 - b - r
\end{equation}

Those quantities, which we will call BER values (benefit, equivalence, risk) from now onwards, express the size of the effect of using $\mathcal{M}$ instead of $\emptyset$ on the probability scale, simultaneously taking into account chosen definition of practical meaningfulness. The BER values are related to ROC curves \cite{Goncalves2014}. More specifically, when $\delta=0$, the $b$ is the area under the ROC curve (AUROC) associated with $Y_{\mathcal{M}[\mathcal{H}]}$  and $Y_{\emptyset[\mathcal{H}]}$, whereas $r+e$ is equal to the area above the curve \citep{Demidenko2016}. This is not a new idea for testing for difference between two distributions. We refer interested readers to \citep{Wolfe1971,Zhou2008,Newcombe2006part1,Newcombe2006part2,Demidenko2016} for more details about ROC curves, computational techniques for estimating AUROC, and application of the method to discriminate between distributions. What is new in our approach is $\delta$, the threshold of practical significance, which should be set in advance according to the theory and empirical knowledge of the optimization-problem class for which $\mathcal{M}[\mathcal{H}]$ is being developed, and making distinction between $r$ and $e$ values - as opposed to \cite{Demidenko2016} who does not distinguish between them.

Finally, we have to address calculation of $b$, $r$, and $e$ and their interpretation. Let us assume that we have run $\mathcal{M}[\mathcal{H}]$ and $\emptyset[\mathcal{H}]$ on problem instances $\pi_1, \pi_2, \ldots, \pi_l$, repeating evaluation on each instance $n$ times using seeds $s_{ij}$, $i=1,2,\ldots,l$, $j=1,2,\ldots,n$. Let $\mathbf{Y}^{*}_{\mathcal{M}}$ and $\mathbf{Y}^{*}_{\emptyset}$ denote $l\times n$ matrices where the results of $\mathcal{M}[\mathcal{H}]$ and $\emptyset[\mathcal{H}]$ are stored. The obvious way to calculate empirical $b$ value, denoted as $b^*$, is to compare the corresponding entries in the result matrices:
\begin{equation}\label{eq:b_star}
	b^* = \frac{1}{l\times n^2}\sum_{i=1}^{l}\sum_{j=1}^{n}\sum_{k=1}^{n}	I_{\mathbf{Y}^{*}_{\mathcal{M}}[i,j] < \mathbf{Y}^{*}_{\emptyset}[i,k] - \delta}
\end{equation}
where $I_\varphi$ is the indicator function that takes the value $1$ when its underlying condition $\varphi$ evaluates to $\top$, and $0$ otherwise.

The explanation of Equation \ref{eq:b_star} is as follows. The result of the sums in Equation \ref{eq:b_star} is equal to the number of times that $\mathcal{M}[\mathcal{H}]$ produced better solutions than $\emptyset[\mathcal{H}]$ for the problems $\pi_1, \pi_2,\ldots,\pi_l$. The denominator $l\times n^2$ is the total number of comparisons.  Therefore, their ratio is an estimate of the probability that for a random instance from the problem class to which $\pi_1,\pi_2,\ldots,\pi_l$  belong, $\mathcal{M}[\mathcal{H}]$ will produce a better solution than $\emptyset[\mathcal{H}]$. Better in this context means "lower for at least $\delta$". 

The empirical equivalence and risk are calculated analogously:

\begin{equation}\label{eq:r_star}
r^* = \frac{1}{l\times n^2}\sum_{i=1}^{l}\sum_{j=1}^{n}\sum_{k=1}^{n}	I_{\mathbf{Y}^{*}_{\mathcal{M}}[i,j] > \mathbf{Y}^{*}_{\emptyset}[i,k] + \delta}
\end{equation}

\begin{equation}\label{eq:e_star}
e^* = \frac{1}{l\times n^2}\sum_{i=1}^{l}\sum_{j=1}^{n}\sum_{k=1}^{n}	I_{\mathbf{Y}^{*}_{\emptyset}[i,k] - \delta < \mathbf{Y}^{*}_{\mathcal{M}}[i,j] < \mathbf{Y}^{*}_{\emptyset}[i,k] + \delta}
\end{equation}

In general, if $\mathcal{M}$ is a good choice for guiding $\mathcal{H}$, we should expect that $b^*>1/2$ and $r^*\approx 0$. If $e^* \approx 1$, then no meaningful difference between $\mathcal{M}[\mathcal{H}]$ and $\emptyset[\mathcal{H}]$ has been found, suggesting that the score of $\mathcal{M}[\mathcal{H}]$ is achieved by $\mathcal{H}$. The greater the value of $r^*$, the stronger the evidence that $\mathcal{M}$ guides $\mathcal{H}$ in a way that it deteriorates the effect of the heuristic component. The closer $b^*$ is to $1$, the stronger the evidence in favor of $\mathcal{M}$ being able to efficiently guide $\mathcal{H}$.

Finally, we must stress out that BER values, just as $p$ value, cannot replace a researcher's own reasoning. Just as we should not base conclusions solely on $p$ values being lower or greater than the usual significance thresholds of $0.01$ and $0.05$, we should not regard empirical benefit, risk, and equivalence as a definite answer to the question concerning the examined metaheuristic's efficiency in guiding its heuristic(s). After all, the nature of statistical research is such that only through replications of experiments can a certain hypothesis be accepted or rejected. So, researchers should always plot $\mathbf{Y}^*_{\mathcal{M}}$ and $\mathbf{Y}^*_{\emptyset}$ one against another to visually inspect the empirical distributions. Moreover, similar plots should be made for each problem instance. Only when all that is taken into account, should researchers formulate their conclusions. 

\subsection{Assumptions}\label{subsec:assumptions}
We will conclude this section by briefly stating the assumptions of the methodology which we proposed:
\begin{enumerate}
	\item[A1] The heuristic component $\mathcal{H}$ is known to work well.
	\item[A2] The performance metric $Y$ is univariate (its value is a single score, not a tuple).
	\item[A3] The metric $Y$ is measured for each run.
\end{enumerate}
We can see that they are fairly general and easy to meet in practice. In Section \ref{sec:conclusion}, we discuss the cases of their violation. 

\section{Experimental Example}\label{sec:example}
In this Section, we will describe how we applied the method presented in Section \ref{sec:proposed_methodology} to a variant of Boolean Satisfiability Problem, 3-SAT, Simulated Annealing (SA), and the SAT heuristic known as Flip. The problem and the algorithms are presented in Sections \ref{subsec:boolean_satisfiability_problem}-\ref{subsec:derivation}. We describe benchmarks in Section \ref{subsec:benchmarks}, tuning in Section \ref{subsec:tuning}, and the results of comparing SA[Flip] to $\emptyset$[Flip] in Section \ref{subsec:results}.

The repository with the code and data can be downloaded from \url{https://osf.io/f2m9w/}. . 
\subsection{Boolean Satisfiability Problem}\label{subsec:boolean_satisfiability_problem}

Boolean Satisfiability Problem, shorthand SAT, is an NP-complete problem \citep{Cook1971} formulated as follows:
\begin{definition}
	Given a Boolean formula $F$ with $n$ propositional letters $p_1, p_2, \ldots, p_n$, find their valuation under which $F$ evaluates to $\top$.
\end{definition}
Any Boolean formula can be converted to conjunctive normal form (CNF), i.e. a conjunction of clauses that are themselves disjunctions of literals (propositional letters or their negations):
\begin{equation}
	\bigwedge_{i=1}^{m}\bigvee_{j=1}^{k_i}\pm p_{i,j}
\end{equation}
where $\pm p_{i,j} \in \{p_l, \neg p_l\}$ for some $l \in \{1,2,\ldots,n\}$. If all $k_i=k$, then we say that $F$ is in its $k$-CNF and refer to the SAT problem as $k$-SAT. Since $3$-SAT is also NP-complete and all Boolean formulae can be converted to $3$-CNF \citep{Cook1971}, we will focus on the case where $k=3$.

\subsection{Solution Representation and Objective Function}\label{subsec:solution_representation_and_objective_function}

A solution to a ($3$-)SAT instance is a valuation of its propositional letters $p_1, p_2, \ldots, p_n$, i.e. a mapping from $\{p_1, p_2, \ldots, p_n\}$ to $\{\bot, \top\}$. Encoding $\bot$ as $0$ and $\top$ as $1$, we can represent solutions as integer arrays of zeros and ones. The goal is to find such a solution that all the clauses in the formula are satisfied. We can formulate the objective function as the percentage of satisfied clauses and aim to maximize it, or the percentage of unsatisfied clauses and try to minimize it. The two objective functions are equivalent and both return the values between $0$ and $1$. We opt for the minimization alternative in this paper (i.e. we will minimize the ratio of the number of unsatisfied clauses to the total number of clauses) and use it as the performance metric $Y$. Obviously, if $Y=0$, the optimal solution has been found and satisfiability of the formula in question has been proven. The reason why we use percentages rather than numbers of unsatisfied clauses is that we want the performance metric to be on the same scale for all formulae, no matter how much clauses they consist of.

\subsection{Flip Heuristic}\label{subsec:flip}
Let $F$ be a $3$-CNF formula. The Flip heuristic receives a possible solution $\mathbf{x}=[x_1,x_2,\ldots,x_n]$ ($x_i \in \{0,1\}$ for $i=1,2,\ldots,n$) and iteratively flips one its element at a time if it improves the objective function until no further improvement is possible. The heuristic is presented in Algorithm \ref{alg:flip_heuristic} \citep{Marchiori1999}.

\begin{algorithm}
	\caption{The Flip Heuristic}
	\label{alg:flip_heuristic}
	\begin{algorithmic}[1]
		\Require $\mathbf{x} = [x_1, x_2, \ldots, x_n]$ - valuation to improve
		\Ensure $\mathbf{x}' = [x'_1, x'_2,\ldots,x'_n]$ - a possibly improved version of $\mathbf{x}$.
		\State $\mathbf{x}' \leftarrow $ copy $\mathbf{x}$
		\State $S \leftarrow $ a random permutation of $[1,2,\ldots,n]$
		\State $improvement \leftarrow 1$
		\While{$improvement > 0$}
			\State $improvement \leftarrow 0$
			\For{$i \leftarrow 1, 2, \ldots n$}
				\State $j \leftarrow S[i]$
				\State $x'_j \leftarrow$ flip $x'_j$
				\State $gain \leftarrow $ difference in number of unsatisfied clauses before and after the flip
				\If{$gain \geq 0$}
					\State $improvement \leftarrow improvement + gain$
				\Else
					\State Reverse the flip
				\EndIf
			\EndFor
		\EndWhile
		\State \Return $\mathbf{x}'$ 
	\end{algorithmic}
\end{algorithm}

\subsection{Simulated Annealing}\label{subsec:simulated_annealing}
Simulated Annealing (SA) is a well-known and widely used metaheuristic whose history dates back to 1980s when  \cite{Kirkpatrick1983} and \cite{Cerny1985} published first papers on the algorithm. Simulated Annealing is inspired by the process of physical annealing with solids, "in which a crystalline solid is heated and then allowed to cool very slowly until it achieves its most regular possible crystal lattice configuration (i.e., its minimum lattice energy state), and thus is free of crystal defects." \citep{Nikolaev2010}. The algorithm starts with an initial solution and processes it iteratively. Each iteration consists of several steps, and at each step, the algorithm compares the current solution to one if its neighbors. The current solution is always replaced with the better neighbor. If the neighbor is worse, replacements occur with a probability which depends on the current temperature. The algorithm receives the initial temperature and the cooling schedule at the beginning and decreases temperature at each iteration according to the schedule. The pseudo-code of the Simulated Annealing is outlined in Algorithm \ref{alg:simulated_annealing} \citep{Nikolaev2010}.

\begin{algorithm}
	\caption{Simulated Annealing - The General Form}
	\label{alg:simulated_annealing}
	\begin{algorithmic}[1]
		\Require $T_0$ - initial temperature, $C : \mathbb{R} \rightarrow \mathbb{R}$ - cooling schedule, $E$ - the objective function to minimize, $M$ - number of steps in an iteration
		\Ensure $\mathbf{x}$ - a solution that should minimize $E$.
		\State $\mathbf{x} \leftarrow $ generate a random solution to start with
		\State $k \leftarrow 0$
		\While{stopping criterion is not met}
			\For{$m \leftarrow 1, 2, \ldots, M$}
				\State $\mathbf{x}' \leftarrow $ randomly generate a neighbor of $\mathbf{x}$
				\State $\Delta E \leftarrow E(\mathbf{x}') - E(\mathbf{x})$
				\If{$\Delta E \leq 0$}
					\State $\mathbf{x} \leftarrow \mathbf{x}'$
				\Else
					\State $\mathbf{x} \leftarrow \mathbf{x}'$ with probability $\exp\left(-\Delta E / T_k\right)$
				\EndIf
			\EndFor
			\State $T_{k+1} \leftarrow C\left(T_k\right)$
			\State $k \leftarrow k + 1$
		\EndWhile
		
		\State \Return $\mathbf{x}$ 
	\end{algorithmic}
\end{algorithm}

\subsection{The SA[Flip] Algorithm for the SAT problem}\label{subsec:flipsa}

The combination of Simulated Annealing and the Flip heuristic for SAT, named SA[Flip], is presented in Algorithm \ref{alg:flip_sa}. In it, the heuristic specific to SAT is applied to the initial solution at the beginning of the algorithm, and once to each neighbor proposed to replace the current solution. Even though there may be other ways to combine the two algorithms, the goal of our study is not to find the best combination of them all, but to show how we can assess if the overall result of the combination being examined is due to the heuristic alone. The same procedure can be carried out for any metaheuristic and the heuristic(s) it guides.

We used geometric cooling schedule, in which $T_{k+1} = \alpha T_k$ for some constant $\alpha \in (0, 1)$. As the stopping criterion we used the following compound condition:
\begin{itemize}
	\item The objective function (ratio of the number of unsatisfied clauses to the total number of clauses) of the current solution is equal to $0$ or
	\item  $k$, the number of iterations performed, is equal to $MNI$, the maximal number of iterations allowed, specified as a SA[Flip]'s parameter.
\end{itemize}
We checked for the stopping condition at each iteration as well as after each step.

Also, we kept track of the best solution encountered during execution of the algorithm and output it when SA[Flip] stops. We decided to do so because it may happen that the algorithm finds the optimal solution, but replaces it with a neighbor that is worse than it.

Even though definition of neighborhoods can be treated as an additional parameter to calibrate, we chose not do so, but to adopt one neighborhood definition in advance in order to reduce the number of parameters and simplify demonstration of our methodology. Of course, we advise researchers to experimentally determine the best definition of a neighborhood, as in \citep{Simic2017}. The one that we adopted and used throughout the experiment is as follows:
\begin{definition}
	Two solutions to the same instance of $3$-SAT problem are neighbors to each other if and only if their Hamming distance is equal to $1$.
\end{definition}
This means that a neighbor of a solution differs from it in valuation of a single propositional letter.

\begin{algorithm}
	\caption{The SA[Flip] algorithm for Boolean SAT problem}
	\label{alg:flip_sa}
	\begin{algorithmic}[1]
		\Require $T_0$ - initial temperature, $\alpha$ - cooling constant, $M$ - number of steps in an iteration, $MNI$ - maximal number of iterations
		\Ensure $\mathbf{x}^{bfs}$ - the best found solution.
		\State $\mathbf{x} \leftarrow $ generate a random valuation to start with
		\State $\mathbf{x} \leftarrow $ apply Flip to $\mathbf{x}$
		\State $\mathbf{x}^{bfs} \leftarrow \mathbf{x}$
		\State $k \leftarrow 0$
		\While{$k < MNI$}
		\For{$m \leftarrow 1, 2, \ldots, M$}
		\State $\mathbf{x}' \leftarrow $ randomly generate a neighbor of $\mathbf{x}$
		\State $\mathbf{x}' \leftarrow $ apply Flip to $\mathbf{x}'$
		\If{$Y(\mathbf{x}') = 0$}
			\State \Return $\mathbf{x}'$
		\EndIf
		\If{$Y(\mathbf{x}) < Y\left(\mathbf{x}^{bfs}\right)$}
			\State $\mathbf{x}^{bfs} \leftarrow \mathbf{x}$
		\EndIf
		\State $\Delta Y \leftarrow Y(\mathbf{x}') - Y(\mathbf{x})$
		\If{$\Delta Y \leq 0$}
		\State $\mathbf{x} \leftarrow \mathbf{x}'$
		\Else
		\State $\mathbf{x} \leftarrow \mathbf{x}'$ with probability $\exp\left(-\Delta Y / T_k\right)$
		\EndIf
		\EndFor
		\State $T_{k+1} \leftarrow \alpha T_k$
		\State $k \leftarrow k + 1$
		\EndWhile
		
		\State \Return $\mathbf{x}^{bfs}$ 
	\end{algorithmic}
\end{algorithm}

\subsection{Derivation of $\emptyset$[Flip]}\label{subsec:derivation}
As said in Section \ref{sec:proposed_methodology}, we need to compare SA[Flip] to $\emptyset$[Flip] in order to estimate how good SA is at guiding Flip. We do not need to explicitly state $\emptyset$ as an actual algorithm. It is sufficient to remove all SA's components from SA[Flip] and leave only naive operations at the metaheuristic level: random generation of the initial solution, random generation of neighbors, and their random acceptance. The parameters inherited from SA[Flip] are $MNI$ and $M$ and they should be set to the same values as for SA[Flip] in order to ensure that $\emptyset$[Flip] can invest the same computational effort as SA[Flip].

We present $\emptyset$[Flip] in Algorithm \ref{alg:naive_flip}.

\begin{algorithm}
	\caption{Algorithm $\emptyset$[Flip], derived from SA[Flip], for Boolean SAT problem}
	\label{alg:naive_flip}
	\begin{algorithmic}[1]
		\Require $M$ - number of steps in an iteration, $MNI$ - maximal number of iterations
		\Ensure $\mathbf{x}^{bfs}$ - the best found solution.
		\State $\mathbf{x} \leftarrow $ generate a random valuation to start with
		\State $\mathbf{x} \leftarrow $ apply Flip to $\mathbf{x}$
		\State $\mathbf{x}^{bfs} \leftarrow \mathbf{x}$
		\State $k \leftarrow 0$
		\While{$k < MNI$}
		\For{$m \leftarrow 1, 2, \ldots, M$}
		\State $\mathbf{x}' \leftarrow $ randomly generate a neighbor of $\mathbf{x}$
		\State $\mathbf{x}' \leftarrow $ apply Flip to $\mathbf{x}'$
		\If{$Y(\mathbf{x}') = 0$}
		\State \Return $\mathbf{x}'$
		\EndIf
		\If{$Y(\mathbf{x}) < Y\left(\mathbf{x}^{bfs}\right)$}
		\State $\mathbf{x}^{bfs} \leftarrow \mathbf{x}$
		\EndIf
		\State $\mathbf{x} \leftarrow \mathbf{x}'$ with random probability $p \in [0, 1]$
		\EndFor
		\State $k \leftarrow k + 1$
		\EndWhile
		
		\State \Return $\mathbf{x}^{bfs}$ 
	\end{algorithmic}
\end{algorithm}

\subsection{Benchmarks}\label{subsec:benchmarks}
Even though $3$-SAT constitutes a class of problems of its own, we did not aim to cover all the possible subclasses of $3$-SAT problems. Instead, we focused on those $3$-SAT instances which are in the so called phase transition. Those are the formulae with approximately $4.24n$ clauses \citep{Gent1994}, where $n$ is the number of propositional letters that appear in them. Such instances are computationally hardest to solve and the probability of them being satisfiable is approximately equal to the probability that they are not.

We also limited $n$ since it is impossible to conduct an experiment involving all possible numbers of propositional letters and our computational resources were limited. We chose the range $50 \leq n \leq n$ because the corresponding solution spaces are sufficiently large but not too much for our testing machine. For $n=50, 75, 100, 125$, we downloaded $100$ corresponding instances from SATLIB (\url{http://www.cs.ubc.ca/~hoos/SATLIB/benchm.html}) \citep{Hoos2000}. They are all satisfiable and in the phase transition. We split the formulae into training and test sets. The former contained $100$ formulae, $20$ for each $n=50, 75, 100, 125$, while the latter included the rest.

\subsection{Tuning the Parameters}\label{subsec:tuning}
We tuned the parameters following the methodology of \cite{Simic2017} as it rigorously employs statistical techniques from Design of Experiments \citep{Montgomery2000}.

\begin{figure}
	\subfloat[Main effect of $M$]{\includegraphics[width = 2.5in]{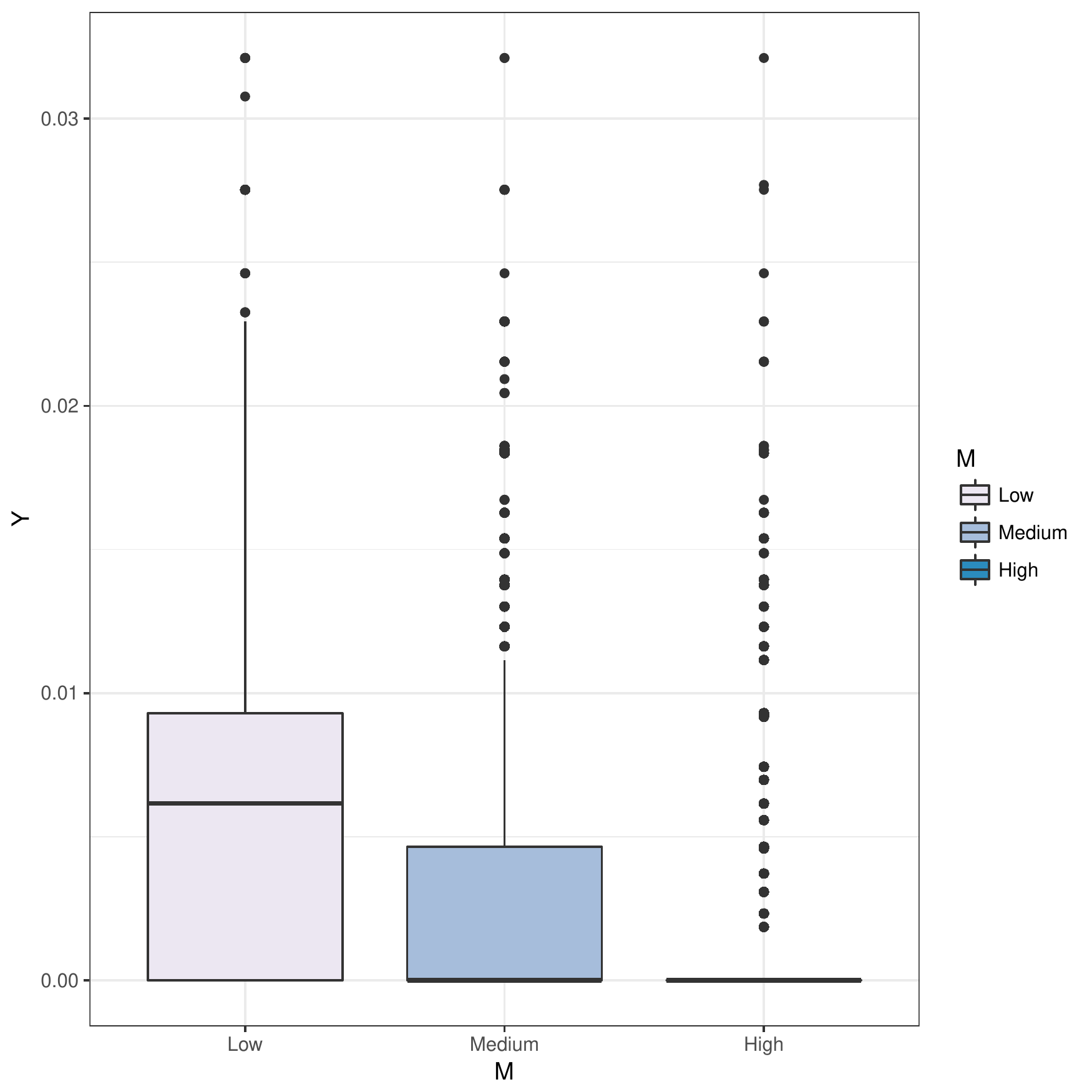}} 
	\subfloat[Main effect of $MNI$]{\includegraphics[width = 2.5in]{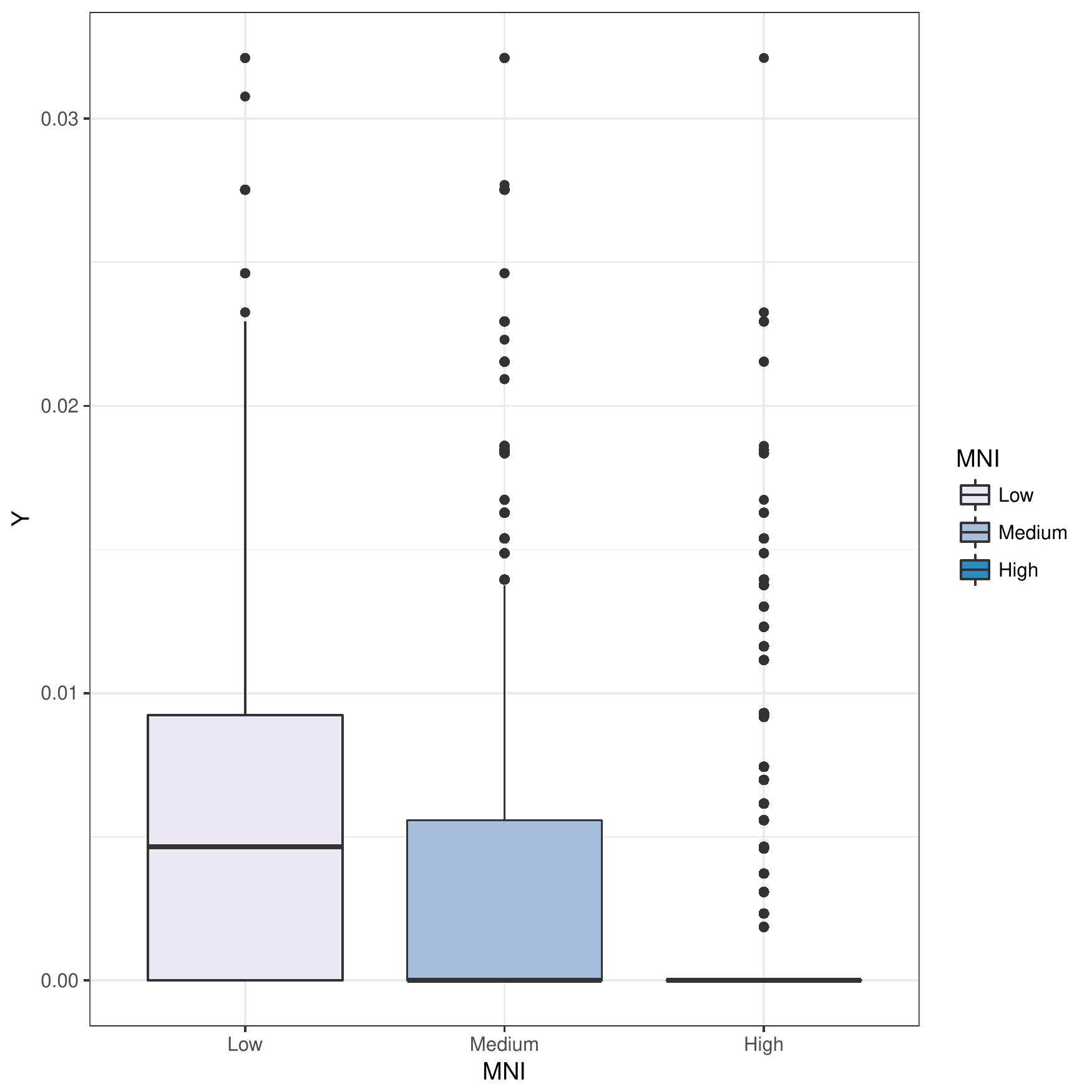}} \\
	\subfloat[Interaction of $M$ and $MNI$]{\includegraphics[width = 2.5in]{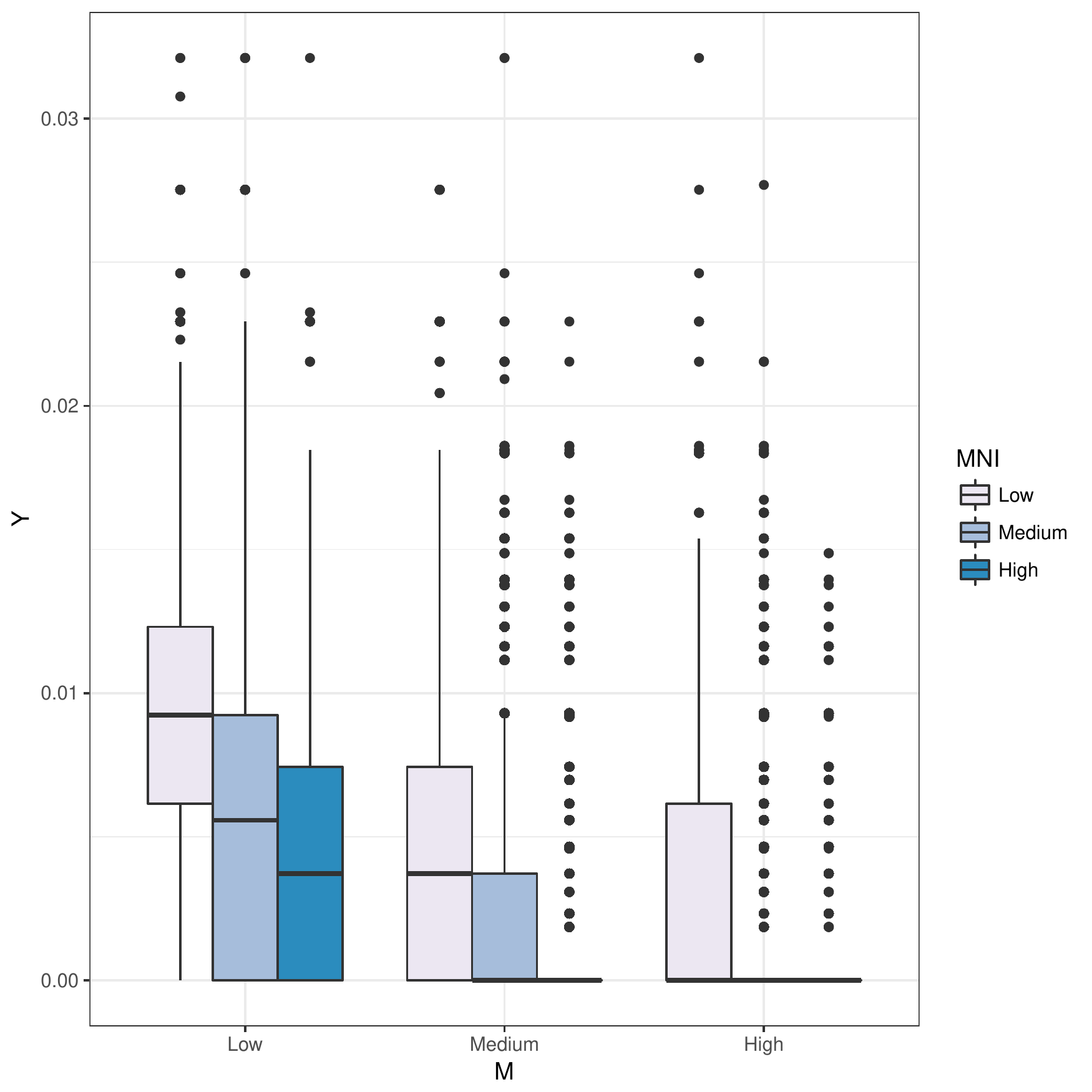}} 
	\subfloat[Interaction of $T_0$ and $\alpha$]{\includegraphics[width = 2.5in]{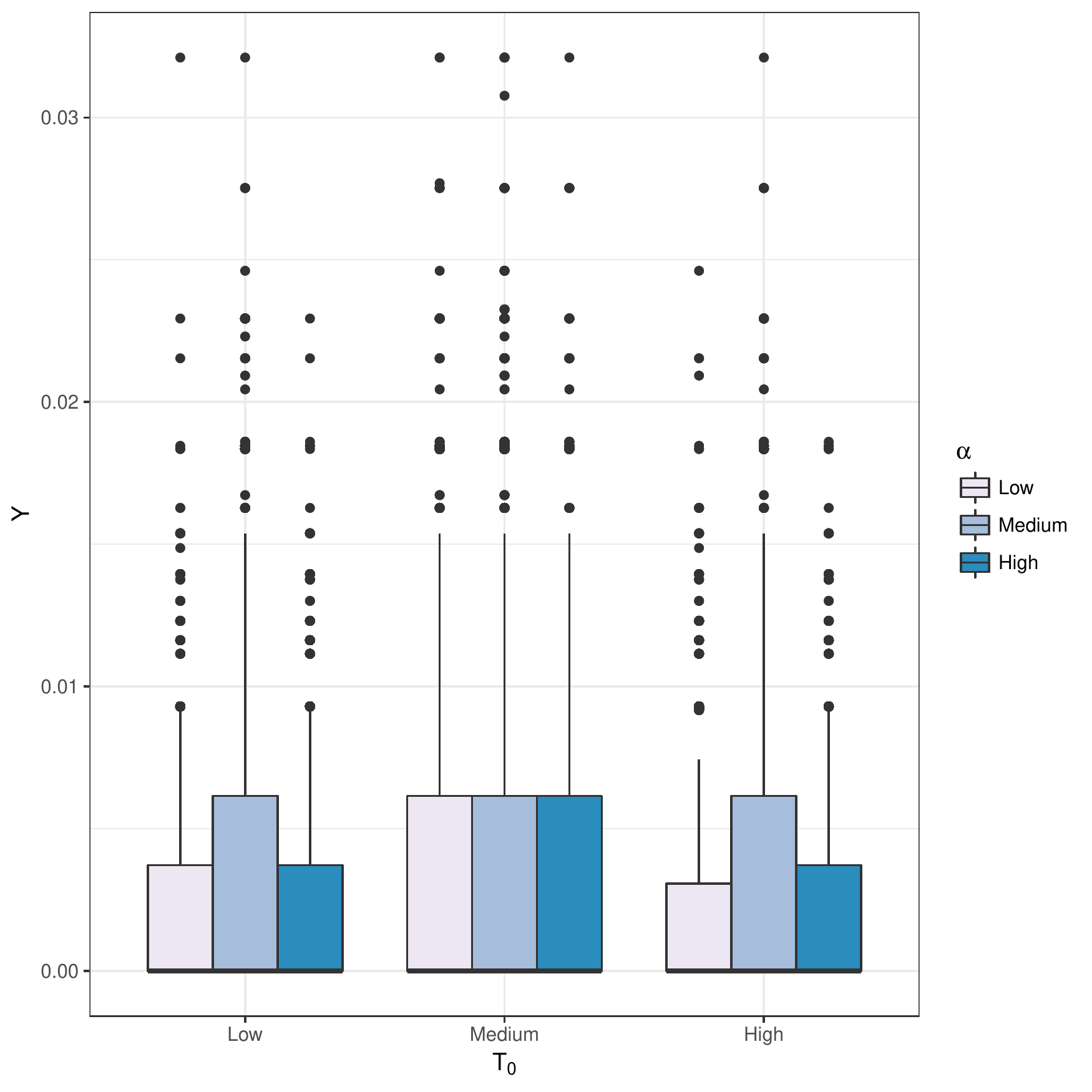}}  
	\caption{Important main effects of the parameters of SA[Flip] and their interactions, identified in the screening phase of the experiment.}
	\label{fig:screening_effects}
\end{figure}

First, we screened the parameters $T_0$, $\alpha$, $M$, and $MNI$ to identify the influential ones. To do so, we defined their low, medium, and high levels (see Table \ref{tab:levels}). We used Box-Behnken design for four three-level factors \citep{Oehlert2000,BoxBehnken1960} and evaluated SA[Flip] for thirty times on each formula in the training set, blocking  the design for seeds. It turned out that $M$ and $MNI$ had substantial main effects and that there were second-order interactions between $T_0$ and $\alpha$, on one hand, and $M$ and $MNI$ on the other. Hence, we had to calibrate all the four parameters. The found effects are presented in Figure \ref{fig:screening_effects}.

Then, we calibrated the parameters iteratively, conducting Response Surface Methodology \#, evaluating SA[Flip] for thirty times on each benchmark, but without blocking the design for seeds. The design that we used in this phase was $2^{4-1}$ fractional factorial. The reason why we used such a simple design is that the response (average performance) can be approximated with a linear model if the portion of the search space is sufficiently small. To ensure that, we used small but effective half-distances for the parameters. We present them in Table \ref{tab:half-distances}. The starting configuration was: $T_0=50$, $\alpha=0.9$, $M=20$ and $MNI=50$, because screening indicated that it might give very good results. We stopped the procedure once the values of $M$ and $MNI$ were such that the maximal number of applications of Flip exceeded $5000$. The found settings are: $T_0=51.71$, $\alpha=0.92$, $M=50$ and $MNI=103$.

\begin{table}
	\centering
	\caption{ Low, medium, and high levels of SA[Flip]'s parameters.  }
	\label{tab:levels}
	\begin{tabular}{llll}
		\hline\noalign{\smallskip}
		Parameter & Low & Medium & High \\
		\noalign{\smallskip}\hline\noalign{\smallskip}
		$T_0$ & $1$ & $10^2$ & $10^3$ \\
		$\alpha$ & $0.5$  & $0.85$ & $0.99$ \\
		$M$ & $1$ & $10$ & $20$ \\
		$MNI$ & $10$ & $50$ & $100$ \\
		\noalign{\smallskip}\hline
	\end{tabular}
\end{table}

\begin{table}
	\centering
	\caption{ Half-distances of the SA[Flip] parameters, used during calibration.  }
	\label{tab:half-distances}
	\begin{tabular}{lllll}
		\hline\noalign{\smallskip}
		Parameter & $T_0$ & $\alpha$ & $M$  & $MNI$\\
		\noalign{\smallskip}\hline\noalign{\smallskip}
		Half-distance & $10$ & $0.04$ & $5$ & $10$ \\
		\noalign{\smallskip}\hline
	\end{tabular}
\end{table}

\subsection{Results and Their Interpretation}\label{subsec:results}

We evaluated SA[Flip] and $\emptyset$[Flip] on each testing benchmarks thirty times. We made sure that the algorithms used the same sets of seeds for each formula to allow for fair comparison.

In general, both methods achieved very good results, as can be seen in Tables \ref{tab:results_grouped_by_n_and_algorithm}-\ref{tab:success_rates_grouped_by_algorithm}. Their performance scores, $Y$, deteriorate as $n$ increases, which is also the case with their success rates - the percentages of successful runs, i.e. the runs where the output $Y=0$.

\begin{table}
	\centering
	\caption{ Average percentages of unsatisfied clauses ($Y$) in Testing Formulae with different numbers of propositional letters $n$, for SA[Flip] and $\emptyset$[Flip]. }
	\label{tab:results_grouped_by_n_and_algorithm}
	\begin{tabular}{lll}
		\hline\noalign{\smallskip}
		$n$ & $\emptyset$[Flip] & SA[Flip]  \\
		\noalign{\smallskip}\hline\noalign{\smallskip}
		$50$ & $0.000000e+00$ & $0.000000e+00$ \\
		$75$ & $5.256542e-05$ & $6.154000e-05$ \\
		$100$ & $2.587233e-04$ & $2.383738e-04$ \\
		$125$ & $5.583783e-04$ & $5.847133e-04$ \\
		\noalign{\smallskip}\hline
	\end{tabular}
\end{table}

\begin{table}
	\centering
	\caption{ Average Percentages of Unsatisfied Clauses ($Y$) in Testing Formulae for SA[Flip] and $\emptyset$[Flip]. }
	\label{tab:results_grouped_by_algorithm}
	\begin{tabular}{ll}
		\hline\noalign{\smallskip}
		Algorithm & Average $Y$ \\
		\noalign{\smallskip}\hline\noalign{\smallskip}
		 $\emptyset$[Flip] & $0.0002174168$ \\
		 SA[Flip] & $0.0002211568$ \\
		\noalign{\smallskip}\hline
	\end{tabular}
\end{table}

\begin{table}
	\centering
	\caption{ Success rates of SA[Flip] and $\emptyset$[Flip] for Testing Benchmarks with different number of propositional leters, $n$. }
	\label{tab:success_rates_grouped_by_n_and_algorithm}
	\begin{tabular}{lll}
		\hline\noalign{\smallskip}
		$n$ & $\emptyset$[Flip] & SA[Flip] \\
		\noalign{\smallskip}\hline\noalign{\smallskip}
		$50$ & $100.00\%$ & $100.00\%$ \\
		$75$ & $99.33\%$ & $99.12\%$ \\
		$100$ & $96.04\%$ & $96.42\%4$ \\
		$125$ & $89.96\%$ & $89.88\%$\\
		\noalign{\smallskip}\hline
	\end{tabular}
\end{table}

\begin{table}
	\centering
	\caption{ Success rates of SA[Flip] and $\emptyset$[Flip]. }
	\label{tab:success_rates_grouped_by_algorithm}
	\begin{tabular}{ll}
		\hline\noalign{\smallskip}
		Algorithm & Average $Y$ \\
		\noalign{\smallskip}\hline\noalign{\smallskip}
		$\emptyset$[Flip] & $96.33\%$ \\
		SA[Flip] & $96.35\%$ \\
		\noalign{\smallskip}\hline
	\end{tabular}
\end{table}

\begin{figure}
	\subfloat[$\delta=0$]{\includegraphics[width = 2.5in]{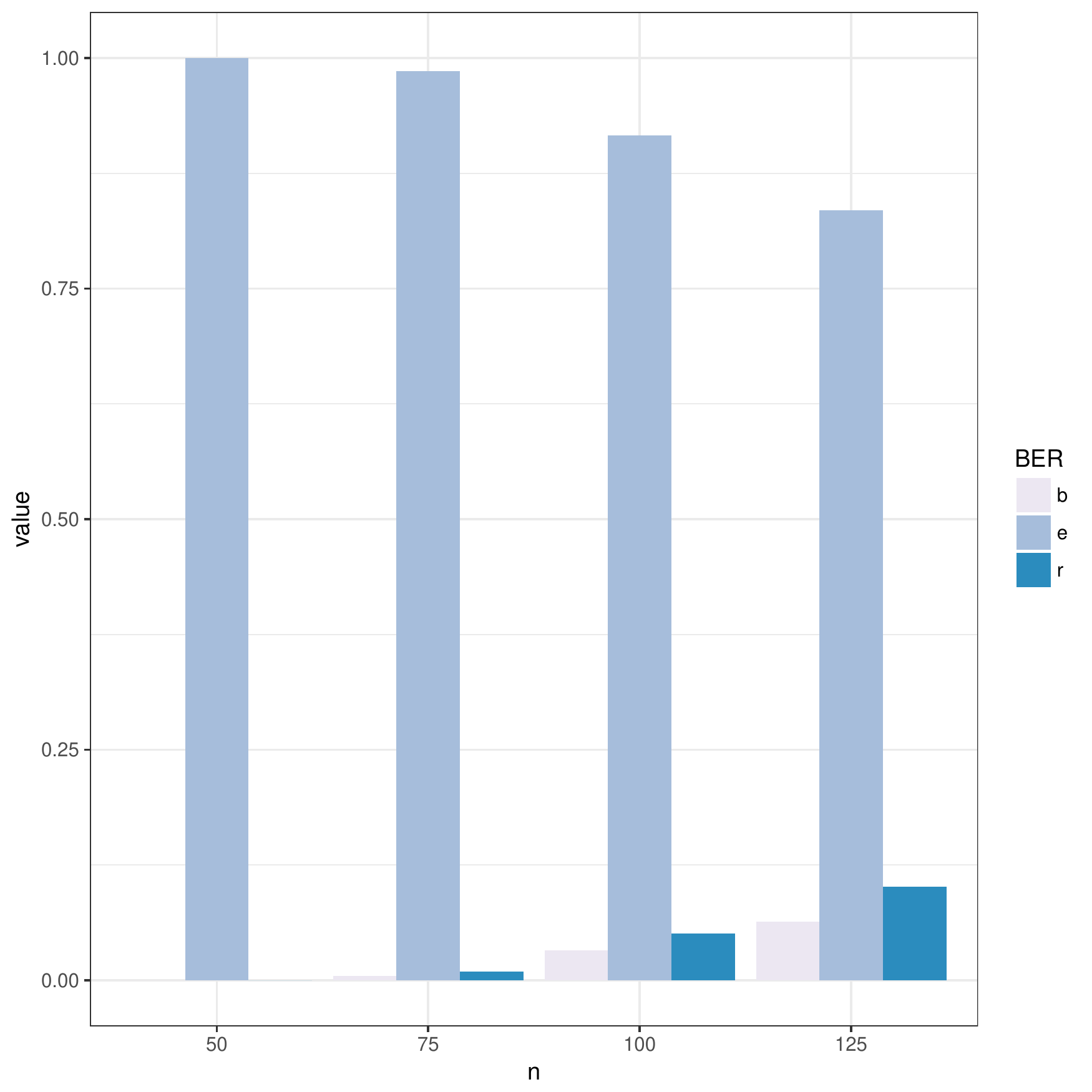}} 
	\subfloat[$\delta=0.01$]{\includegraphics[width = 2.5in]{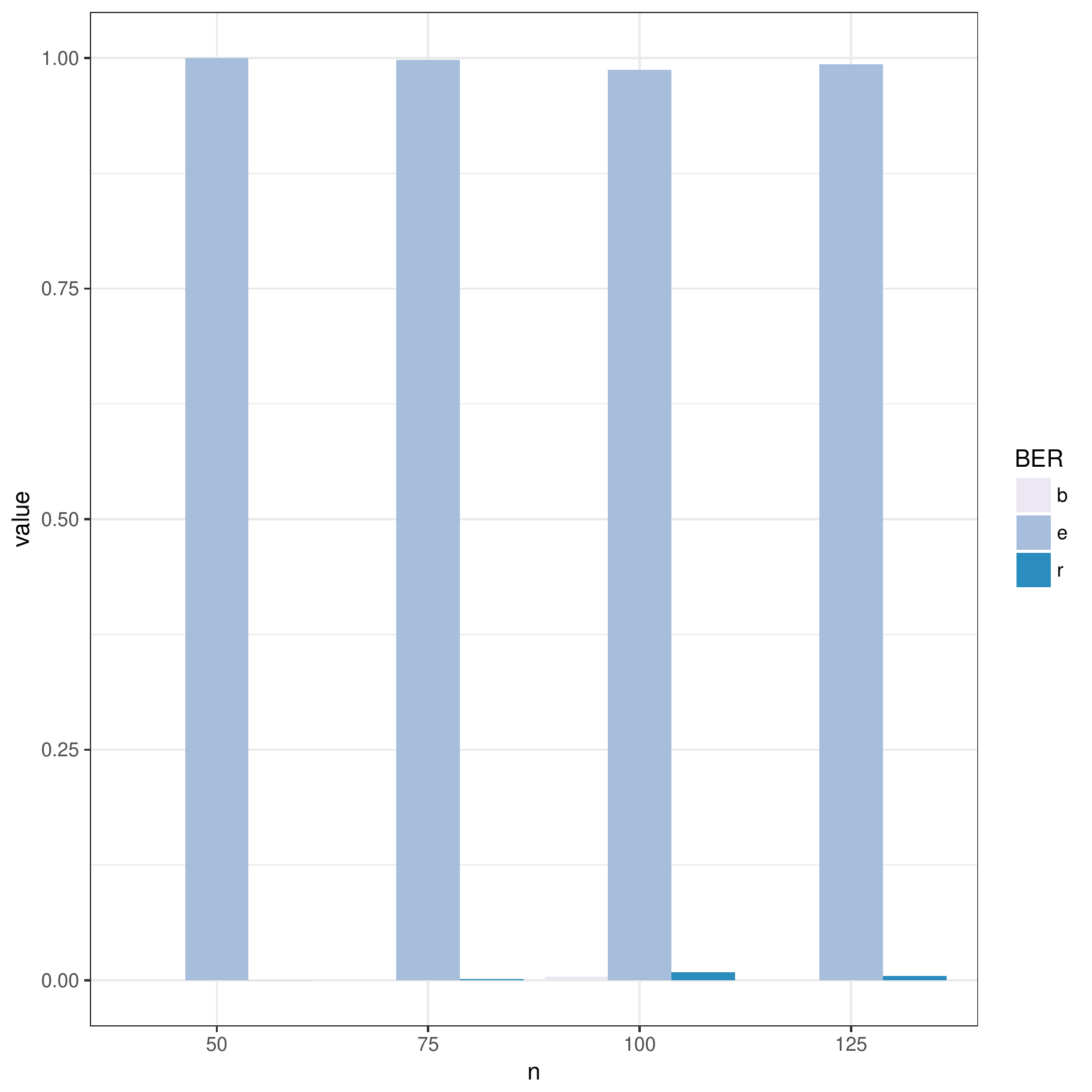}} \\
	\subfloat[$\delta=0.02$]{\includegraphics[width = 2.5in]{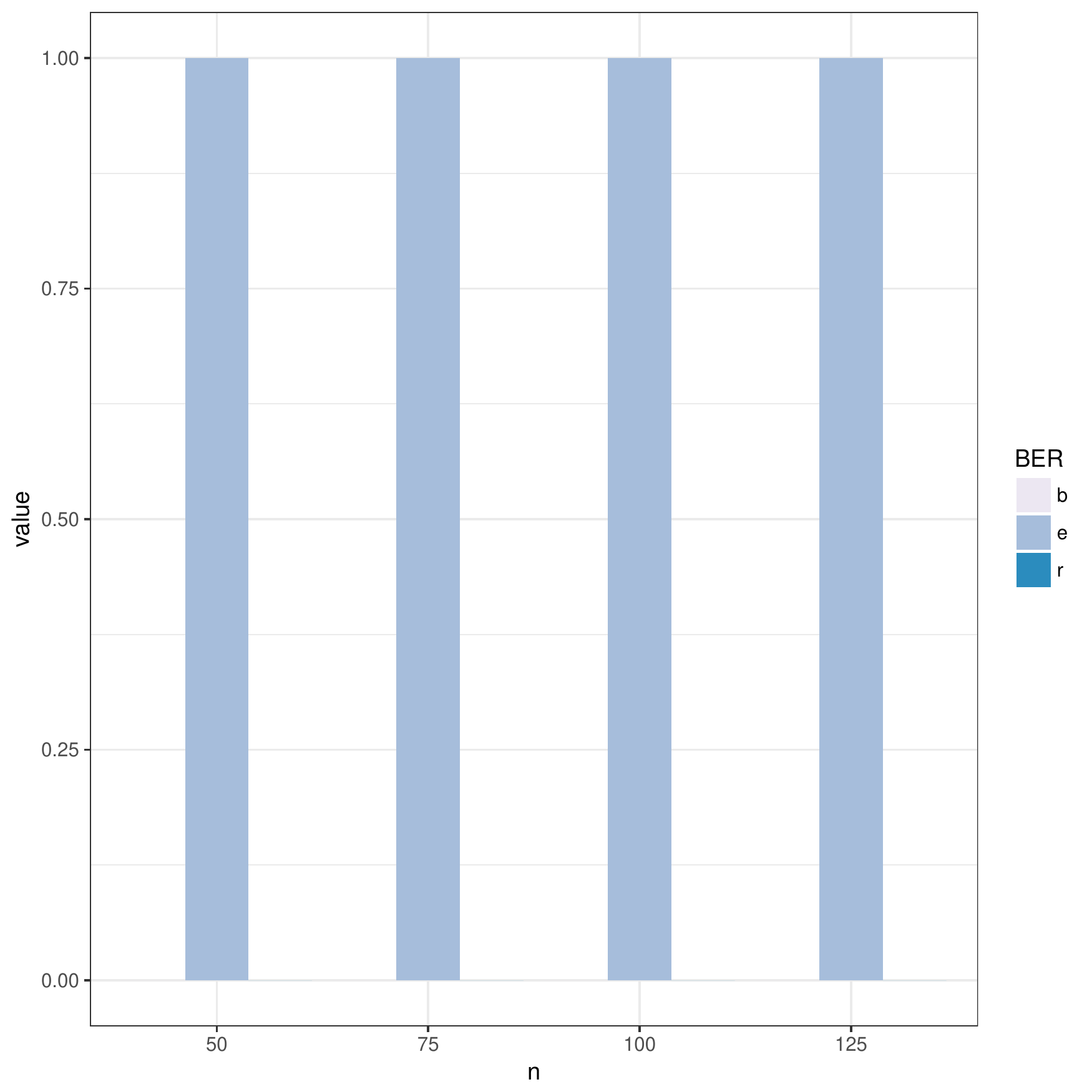}} 
	\caption{BER values for different choices of $\delta$}
	\label{fig:ber_values}
\end{figure}

We calculated BER values for each $n=50,75,100,125$ as well as for the whole test set. We used three different values for $\delta$ in our analysis: $0$, $0.01$, and $0.02$. The results are presented in Tables \ref{tab:ber_values_for_delta_0}-\ref{tab:ber_values_for_delta_0.02} and depicted in Figure \ref{fig:ber_values}.

\begin{table}
	\centering
	\caption{Empirical BER values for $\delta=0.0000$}
	\label{tab:ber_values_for_delta_0}
	\begin{tabular}{llll}
		\noalign{\smallskip}\hline\noalign{\smallskip}
		$n$ & $b^*$ & $e^*$ & $r^*$ \\
		\noalign{\smallskip}\hline\noalign{\smallskip}
		$50$  & $0.0000$  & $1.0000$  & $0.0000$ \\
		$75$  & $0.0051$  & $0.9856$  & $0.0093$ \\
		$100$  & $0.0327$  & $0.9164$  & $0.0509$ \\
		$125$  & $0.0638$  & $0.8349$  & $0.1013$ \\
		\noalign{\smallskip}\hline\noalign{\smallskip}
		overall  & $0.0254$  & $0.9342$  & $0.0404$ \\
		\noalign{\smallskip}\hline
	\end{tabular}
\end{table}

\begin{table}
	\centering
	\caption{Empirical BER values for $\delta=0.0100$}
	\label{tab:ber_values_for_delta_0.01}
	\begin{tabular}{llll}
		\noalign{\smallskip}\hline\noalign{\smallskip}
		$n$ & $b^*$ & $e^*$ & $r^*$ \\
		\noalign{\smallskip}\hline\noalign{\smallskip}
		$50$  & $0.0000$  & $1.0000$  & $0.0000$ \\
		$75$  & $0.0004$  & $0.9982$  & $0.0014$ \\
		$100$  & $0.0043$  & $0.9870$  & $0.0087$ \\
		$125$  & $0.0017$  & $0.9935$  & $0.0048$ \\
		\noalign{\smallskip}\hline\noalign{\smallskip}
		overall  & $0.0016$  & $0.9947$  & $0.0037$ \\
		\noalign{\smallskip}\hline
	\end{tabular}
\end{table}

\begin{table}
	\centering
	\caption{Empirical BER values for $\delta=0.0200$}
	\label{tab:ber_values_for_delta_0.02}
	\begin{tabular}{llll}
		\noalign{\smallskip}\hline\noalign{\smallskip}
		$n$ & $b^*$ & $e^*$ & $r^*$ \\
		\noalign{\smallskip}\hline\noalign{\smallskip}
		$50$  & $0.0000$  & $1.0000$  & $0.0000$ \\
		$75$  & $0.0000$  & $1.0000$  & $0.0000$ \\
		$100$  & $0.0000$  & $1.0000$  & $0.0000$ \\
		$125$  & $0.0000$  & $1.0000$  & $0.0000$ \\
		\noalign{\smallskip}\hline\noalign{\smallskip}
		overall  & $0.0000$  & $1.0000$  & $0.0000$ \\
		\noalign{\smallskip}\hline
	\end{tabular}
\end{table}

Overall, the $e$ value turned out to dominate other two by large margins for all choices of $\delta$ and each $n=50$, $75$, $100$, and $125$. This implies that SA[Flip] is effectively the same as $\emptyset$[Flip], i.e. that SA guides Flip as effectively as the corresponding naive metaheuristic. We can also observe that $e^*$ drops whereas $b^*$ and $r^*$ increase with $n$ for $\delta=0$. It is probable that such a trend continues for $n > 125$ and SA[Flip] and $\emptyset$[Flip] become effectively distinct at some point. Therefore, future research could focus on investigating this hypothesis.

\begin{figure}
	\subfloat[$\text{SA[Flip]}$]{\includegraphics[width = 2.5in]{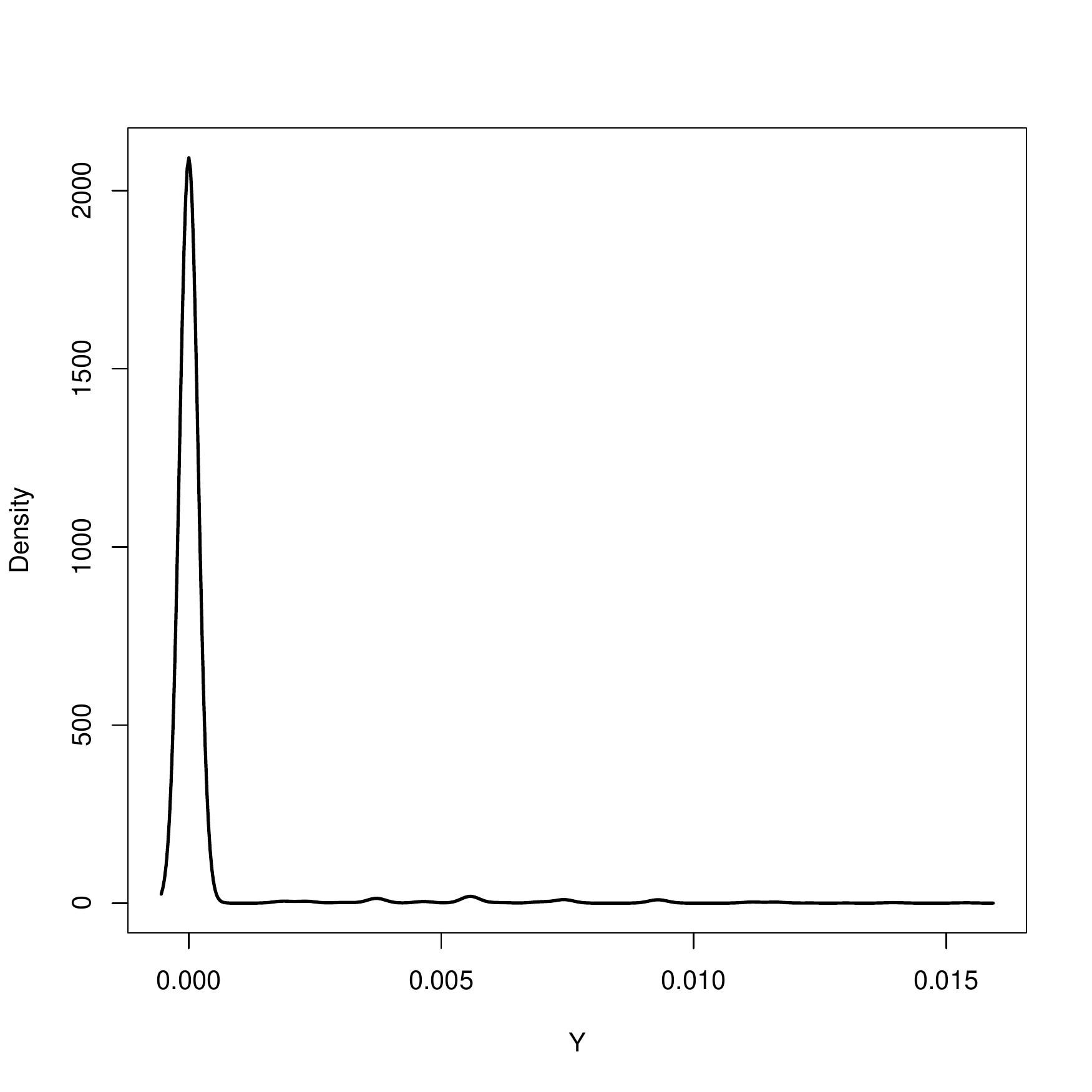}} 
	\subfloat[${\emptyset\text{[Flip]}}$]{\includegraphics[width = 2.5in]{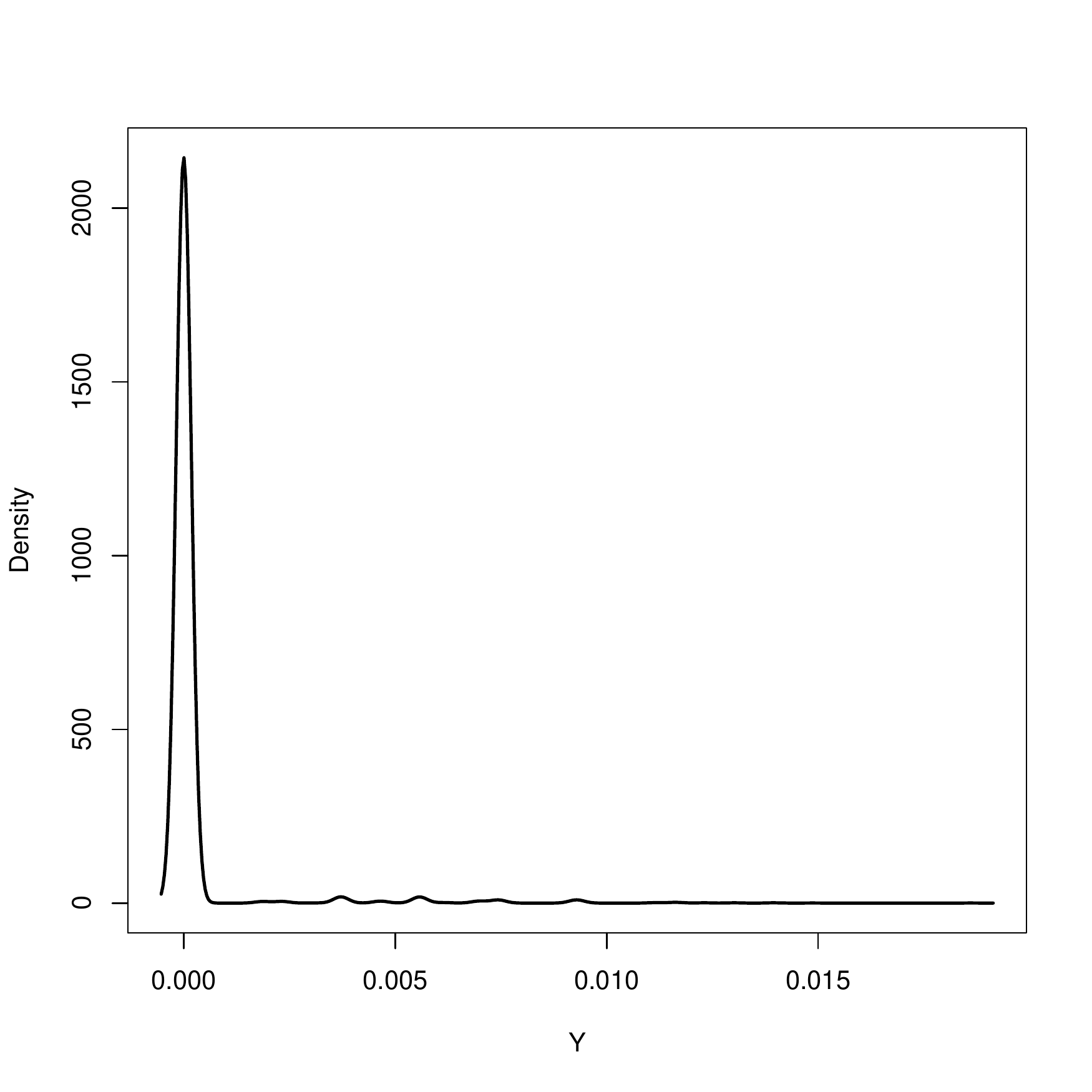}} \\
	\caption{Distributions of performance metric $Y$ for SA[Flip] and $\emptyset$[Flip]}
	\label{fig:plots_of_distributions_one_against_another}
\end{figure}

Plots of distributions of $Y$ for SA[Flip] and $\emptyset$[Flip] are presented in Figure \ref{fig:plots_of_distributions_one_against_another}. By visual inspection, we conclude that the distributions are almost indistinguishable, which confirmes what $e^*$ has indicated: that SA is not than $\emptyset$ at guiding Flip  (for this set of problems). In turn, that means that the observed performance of SA[Flip] is most probably due to Flip alone. Had we not tested SA[Flip] in this manner, we would not have discovered that efficiency of SA[Flip] came from the heuristic component. We were able to find it out only because we compared SA[Flip] to $\emptyset$[Flip], which stresses out the importance of the methodology proposed in this paper and its usefulness in research in this field.

There is one more issue that needs to be discussed. Another explanation for observed equivalence of SA[Flip] and $\emptyset$[Flip] could be that the benchmarks we used are too easy so both algorithms performed really well and no difference between them was possible to found in the first place. This is known as ceiling effect \citep{Bartz2014}. However, we do not think that the effect occurred in this experiment because we deliberately used the benchmarks that are in phase transition and hence, the most challenging and difficult to solve. In addition to this, the numbers of propositional letters were not low and there are studies which evaluated solvers on the same groups of benchmarks but reported worse results, e.g. \citep{Djenouri2016}. Still, researchers who decide to follow our methodology to estimate their metaheuristics should take caution and make sure that their benchmarks do not cause ceiling (or floor) effects.

Finally, our conclusion is as follows. Success of our SA[Flip] for the class of random $3$-SAT formulae in phase transition with $125$ propositional letters at most comes from the Flip heuristic. For this class of Boolean formulae, guiding Flip with Simulated Annealing has the same effect as using a naive random metaheuristic with essentially no logic to guide application of Flip. The more similar SAT problems are to those used in our study, the higher is the probability that the same effects will be detected.

\section{Discussion and Conclusions}\label{sec:conclusion}

In this paper, we have proposed a methodology to empirically estimate how efficient a metaheuristic algorithm $\mathcal{M}$ is at guiding specific heuristic(s) $\mathcal{H}$. The proposed technique was applied to Simulated Annealing (SA), Boolean Satisfiability Problem and the Flip heuristic. The experiment revealed that the performance score of the combination of SA and Flip was due to the heuristic, which is a result that we would not be able to obtain without our methodology. 

The methodology itself is mathematically well-founded, intuitive, easy to apply and relies on practical significance rather than solely on statistics. It directly compares empirical distributions of the chosen performance metrics, not just sample means, which provides a better insight into the metaheuristic in question. By comparing $\mathcal{M}[\mathcal{H}]$ to $\emptyset[\mathcal{H}]$, the technique allows us to estimate the effect of using metaheuristic $\mathcal{M}$ to guide $\mathcal{H}$. The introduced BER values quantify that effect on the probability scale and, accompanied by visual comparison, reveal whether it is justified to guide application of $\mathcal{H}$ with $\mathcal{M}$. Without investigating if performance of $\mathcal{M}[\mathcal{H}]$ comes mostly or entirely from $\mathcal{H}$, we can easily draw wrong conclusions and claim that we have discovered novel solvers, when, if fact,  we have done nothing more than wrapping up an efficient heuristic solver with a metaheuristic whose contribution is negligible. If we propose $\mathcal{M}[\mathcal{H}]$ as a new solver, we must prove that there is something that makes it worth to guide $\mathcal{H}$ with $\mathcal{M}$, and the technique studied and demonstrated in this paper offers a way to do precisely that.

Comparison to $\emptyset[\mathcal{H}]$ is the core of our approach. We defined $\emptyset$ to be a naive, placebo metaheuristic, which performs only naive operations in guiding $\mathcal{H}$. We argued that completely random decisions constitute naive moves and that such $\emptyset$ represents the "low" level of $M$ in Equation \ref{eq:performance_distribution_extended}, equivalent to no guiding logic. Can there be naive operations other than completely random decisions? Can greedy moves be thought of as naive? Those would be operations that always select the best solution from a group of candidates and discard the rest. Although they may seem naive, they do follow some logic, no matter how simple it is. Therefore, $\emptyset[\mathcal{H}]$ that would include greedy moves would actually be a plain greedy algorithm. If $\mathcal{M}[\mathcal{H}]$ fails to beat it, we can say that there are no reasons to use $\mathcal{M}[\mathcal{H}]$ when a simple solver achieves the same or better results, but we would not be able to test if performance of $\mathcal{M}[\mathcal{H}]$ is achieved by $\mathcal{H}$ or the logic of $\mathcal{M}$ contributes to it significantly. Therefore, we argue for $\theta[\mathcal{H}]$ to contain only random operations.

One may also ask why we do not compare $\mathcal{M}[\mathcal{H}]$ to $\mathcal{M}[\emptyset]$, where $\emptyset$ would denote use of no heuristic at all? The reason is that difference between $\mathcal{M}[\mathcal{H}]$ and $\mathcal{M}[\emptyset]$ can reveal if $\mathcal{H}$ contributes anything to the performance of the whole method, not if $\mathcal{M}$ is able to guide it efficiently.

The proposed approach is not without limitations, though. First of all, it requires evaluation of an additional algorithm ($\emptyset[\mathcal{H}]$) which is derived from the metaheuristic being examined. Even though this prolongs research, it also provides information which we would not get otherwise, as demonstrated in the example in Section \ref{sec:example}, and without which scientific conclusions could be flawed. Therefore, we find that taking more time to complete this step pays off. Another limitation is that it enables us to reason about our metaheuristic only with respect to a chosen class of optimization problems. However, this limitation is not unique to this methodology and is inherent to all techniques for analyzing numerical experiments involving stochastic optimization algorithms. 

Also, we must ask if the hybrid algorithm $\mathcal{M}[\mathcal{H}]$ is the only (sensible) combination of $\mathcal{M}$ and $\mathcal{H}$ because if it is the case, then we can be sure that our method sheds light onto general ability of $\mathcal{M}$ to guide $\mathcal{H}$. However, there may be more than one way to guide $\mathcal{H}$ with $\mathcal{M}$. For instance, had we used  Genetic Algorithm (GA) \citep{Holland1992} instead of SA in the example in Section \ref{sec:example}, we could have applied Flip after mutation (as \cite{Marchiori1999}), but we could have also done it before mutation, immediately after performing crossovers. In such cases, rather than estimating general ability of $\mathcal{M}$ to guide $\mathcal{H}$, we are assessing efficiency of the specific strategy based on $\mathcal{M}$ for guiding $\mathcal{H}$. If its effect is approximately equal to that of $\emptyset$, which can be tested with our method, then we can determine if guiding $\mathcal{H}$ in that particular way is justified. Moreover, even though it is possible to plug a heuristic into a metaheuristic between any two operations, is it really sensible to arbitrary intertwine their logics? In each metaheuristic there is a point where the quality of a solution, i.e. the value of the objective function, is computed. It is the only step in execution of metaheuristic methods where they need to evaluate a problem-specific function. All the other operations that they perform are based on specific optimization ideas or some metaphors and constitute a logical unity. In our opinion, the moment just before evaluating a solution is a good time for applying a heuristic because that keeps problem-specific operations (evaluation of the objective function and application of heuristic) at one place, allowing the logic of the metaheuristic to execute without interruptions and as originally designed. This way, comparing $\mathcal{M}[\mathcal{H}]$ to $\emptyset[\mathcal{H}]$ is as close to revealing the general effect of using $\mathcal{M}$ to guide $\mathcal{H}$ as it gets. 

We also need to discuss the assumptions of our methodology as well as the effects of violating them. One of the assumptions is that $\mathcal{H}$ is an efficient heuristic. If $\mathcal{H}$ is new and its efficiency has not been confirmed, then the heuristic must be tested prior to application of the proposed methodology. Another assumption is that performance metric $Y$ is univariate, i.e. a single value, not a tuple of values. If several metrics are of interest, we can compare $\mathcal{M}[\mathcal{H}]$ to $\emptyset[\mathcal{H}]$ once for each of them and then analyze results per metric. Then, what if we want to use a metric calculated as an aggregated value of the results of several runs on each problem instance? Let us suppose that we have stored the results in $l\times n$ matrices $\mathbf{Y}_{\mathcal{M}}^*$ and $\mathbf{Y}_{\emptyset}^*$.  The actual metric scores that we are interested in are then calculated as $Z_{\mathcal{M}}^*[i]=f(\mathbf{Y}^*_{\mathcal{M}}[i,1], \mathbf{Y}^*_{\mathcal{M}}[i,2],\ldots,\mathbf{Y}^*_{\mathcal{M}}[i,n])$ and $Z_{\emptyset}^*[i]=f(\mathbf{Y}^*_{\emptyset}[i,1], \mathbf{Y}^*_{\emptyset}[i,2],\ldots,\mathbf{Y}^*_{\emptyset}[i,n])$ ($i=1,2,\ldots,l$), where $f$ is the aggregating function. The methodology can still be applied, but the formulae for empirical BER values would need to be modified and the results could not be interpreted in quite the same way. Instead of comparing $P(Y | M=\mathcal{M}, \Pi, S, \theta_{\mathcal{M}}, \theta_{\mathcal{H}})$ to $P(Y | M=\mathcal{\emptyset},  \Pi, S, \theta_{\emptyset}, \theta_{\mathcal{H}})$, we would essentially be comparing $P(Z | M=\mathcal{M}, \Pi, \theta_{\mathcal{M}}, \theta_{\mathcal{H}})$ to $P(Z | M=\mathcal{\emptyset},  \Pi, \theta_{\emptyset}, \theta_{\mathcal{H}})$ and the value $b=P(Z_{\mathcal{M}} < Z_{\emptyset} - \delta)$ would answer the following question:
\begin{itemize}
	\item What is the probability that, for a randomly chosen instance from the problem class of interest, $\mathcal{M}[\mathcal{H}]$'s score will be practically better than that of $\emptyset[\mathcal{H}]$ \textbf{when aggregated over several runs}?
\end{itemize}
This is different from the meaning of the $b$ value as originally defined in Section \ref{sec:proposed_methodology} for the non-aggregated case. The corresponding formula for $b^*$ would then be:
\begin{equation}
b^* = \frac{1}{l}\sum_{i=1}^{l}I_{Z_{\mathcal{M}[i]} < Z_{\emptyset}[i] - \delta}
\end{equation}
with analogous modifications being in place for $e^*$ and $r^*$. Those differences are simple, but subtle, so we need to point them out.

Finally, the BER values that we define and propose to quantify the degree to which two distributions are not just statistically, but practically different, can be used to compare any two stochastic algorithms, not just $\mathcal{M}[\mathcal{H}]$ and $\emptyset[\mathcal{H}]$. Moreover, since numerical and practical significance are confounded in BER values through $\delta$, we find them suitable to detect important effects not just in the field of metaheuristics, but in science in general.

We hope that other researchers will see merit in our idea, adopt it in their own studies and improve it further to the benefit of the whole research community.

Possible directions of future research are:
\begin{itemize}
	\item Developing a methodology that would simultaneously test both the efficiency of $\mathcal{H}$ and $\mathcal{M}$'s ability to guide it;
	\item Formulating a technique capable of estimating the general ability of $\mathcal{M}$ to guide any heuristic for the problem at hand, not just the selected $\mathcal{H}$.
\end{itemize}

\clearpage
\bibliography{literature}
\end{document}